\documentclass[10pt,twocolumn,letterpaper]{article}

\usepackage[pagenumbers]{cvpr} 

\usepackage{graphicx}
\usepackage{amsmath}
\usepackage{amssymb}
\usepackage{booktabs}
\usepackage{times}
\usepackage{epsfig}
\usepackage{graphicx}
\usepackage{amsmath}
\usepackage{amssymb}
\usepackage{multirow} 
\usepackage{cancel}
\usepackage{rotating}
\usepackage{latexsym}
\usepackage{adjustbox}
\usepackage[thinlines]{easytable}
\usepackage{subcaption}
\usepackage{booktabs}       
\usepackage{mfirstuc}

\usepackage[pagebackref,breaklinks,colorlinks]{hyperref}

\usepackage[capitalize]{cleveref}
\crefname{section}{Sec.}{Secs.}
\Crefname{section}{Section}{Sections}
\Crefname{table}{Table}{Tables}
\crefname{table}{Tab.}{Tabs.}

\def\cvprPaperID{2891} 
\def\confName{CVPR}
\def\confYear{2022}

\begin{document}

\title{A Framework for Unsupervised Robustness via Sensitive Factor Discovery and Augmentation, Coherence and Adversarial (ACAI) Interventions}

\title{Adaptation and Generalization for Unknown Sensitive Factors of Variations}

\author{William Paul\\
Johns Hopkins University Applied Physics Laboratory\\
11100 Johns Hopkins Road Laurel, Maryland 20723\\
{\tt\small william.paul@jhuapl.edu}
\and
Philippe Burlina\\
Johns Hopkins University Applied Physics Laboratory\\
11100 Johns Hopkins Road Laurel, Maryland 20723\\
{\tt\small philippe.burlina@jhuapl.edu}
}
\maketitle

\begin{abstract}
   Assured AI in  unrestricted settings is a critical problem. Our framework addresses  AI assurance challenges lying at the intersection of domain adaptation, fairness, and counterfactuals analysis, operating via the discovery and intervention on  factors of variations in data (e.g. weather or illumination conditions) that significantly affect the robustness of AI models. Robustness is understood here as insensitivity of the model performance to variations in sensitive factors. 
  
 Sensitive factors are traditionally set in  a supervised setting, whereby factors are known a-priori (e.g. for fairness this could be factors like sex or race). In contrast, our motivation is real-life scenarios where less, or nothing, is actually known a-priori about certain factors that cause models to fail. This leads us to consider various settings (unsupervised, domain generalization, semi-supervised) that correspond to different degrees of incomplete knowledge about those factors. 
   
 Therefore, our two step approach works by  a) discovering  sensitive factors that cause AI systems to fail in a unsupervised fashion, and then b)  intervening models to lessen these factor's influence. 
Our method considers 3 interventions consisting of Augmentation, Coherence, and Adversarial Interventions (ACAI). 
We  demonstrate the ability for interventions on discovered/source factors to generalize to target/real factors. We also demonstrate how adaptation to real factors of variations can be performed in the semi-supervised case where some target factor labels are known, via automated intervention selection. Experiments  show that our approach  improves on baseline models, with regard to achieving optimal utility vs. sensitivity/robustness tradeoffs.

\end{abstract}

\section{Introduction}
Deploying artificial intelligence (AI) systems in real world settings requires greater assurances for robustness and trust in system behavior. The currently held adage inside and outside the AI community is that, while deep learning (DL)  applied to tasks ranging from face recognition 
to medical diagnostics
has reached human performance,
susceptibility to issues including private information leakage \cite{shokri2017membership}
adversarial attacks \cite{carlini2017adversarial}
AI fairness vis-a-vis protected subpopulations, 
remains a key impediment to production- or clinical-grade deployments.

Consider specifically the challenge of fairness, which has recently received significant attention in the literature and press. Fairness focuses on how models may  treat certain sub-populations or individuals adversely compared to others. How exactly this adverse treatment manifests itself depends heavily on context and tasks, but much of the focus of interventions are on addressing disparities in  performance of  discriminative models vis-a-vis different  protected subpopulations defined by factors such as sex or skin color. 

\begin{figure}[t]
\centering 
\includegraphics[page=1,width=\linewidth]{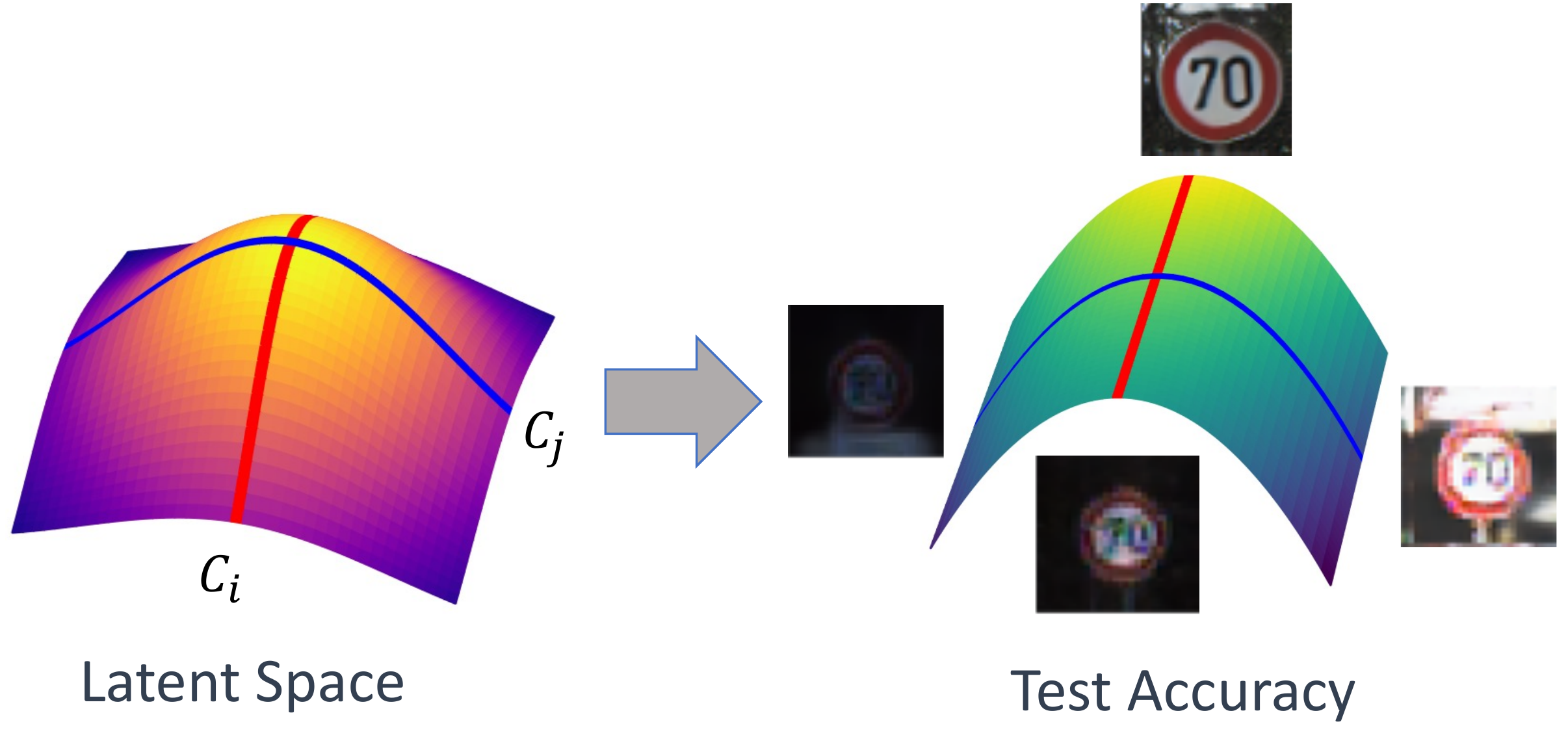}
     \vspace{-0.2in}
    \caption{Main idea: We learn variables such as $C_i$ or $C_j$ in latent space that corresponds to a) factors of variations such as lighting (blue) or sign size (red) and b) to probe or address how sensitive task performance (low performance is purple versus high performance is yellow-green) is to these factors. 
    }
\label{fig:overview1}
\end{figure}
Therefore, a  natural extension of fairness/bias considerations, addressing  more broadly an aspect of robustness of models related to {\em invariance}, 
consists on requiring that models' performance be invariant and insensitive to known factors of variation in the data, so as to ensure consistent/guaranteed levels of performance: e.g., it is desirable to have a pedestrian detector perform with equally well irrespective of weather conditions, or a face verification irrespective of ethnicity. 

In this context, two impediments are notable: 1) {\em Cost constraints}: Collecting/curating more data is key to evaluate and mitigate this type of bias, but is expensive, and sometimes impossible. 
2) {\em Knowledge limitations}: Which factors of variation in the data impact performance is  often not known a-priori; often, not all factors are known; sometimes, if some factors are known, labels for these may be unavailable or scarce. All these represent various knowledge limitations and the need to consider various unsupervised or generalization settings. 

Our response to those issues is to develop a dual-pronged framework that leverages generative models that both create synthetic data and discovers in an unsupervised fashion sensitive factors of variation, and then intervenes models vis-a-vis those factors, for various un/semi-supervised/generalization settings. 


\section{Prior Work} 
Conceptually, our approach generalizes  AI fairness work,
but also stands at the confluence of recent efforts in counterfactual analysis, domain adaptation, and work in generative models.
We review prior state of the art (SOTA) relative to these problems:

{\bf AI Fairness:} Studies have spanned  applications ranging from adult income~\cite{zemel2013learning}, natural language processing~\cite{bolukbasi2016man}, categorical data~\cite{zemel2013learning,prost2019toward}, to skin diagnostics~\cite{kinyanjui2020fairness}. All fairness studies (known to us) are setup for the supervised case where it is assumed that factors are known and factors' labels are available. Here we fully relax that  assumption.

High level taxonomies of interventions  mitigating bias broadly gravitate towards three  categories: (1) {\bf pre-processing} interventions on the training data;
(2) {\bf in-processing}, model-altering interventions;
and (3) {\bf post-processing}  interventions on prediction outputs. 

{\bf Pre-Processing Interventions} alter training data to mitigate data imbalance or correlations between sensitive factors and target labels, and rely broadly on approaches that perform either data re-weighting, or  augmentation, done via generative models or style transfer. Generative models typically used for such augmentations for debiasing or domain adaptation, include variational autoencoders (VAEs)~\cite{madras2018learning,kingma2013auto, louizos2015variational}, GANs~\cite{karras2019style,grover2019bias}, or  flow-based models \cite{zhao2017infovae};   
style transfer include,
e.g., CycleGAN~\cite{cyclegan2017}, StarGAN~\cite{stargan2018} and pix2pix~\cite{pix2pix2017, pix2pix2019}. Examples of prior work on fairness using the above techniques include~\cite{quadrianto2019discovering,hwang2020fairfacegan,sattigeri2018fairness}. However, those approaches lack in two important ways, since they a) leverage generative models fundamentally not designed to do fine control and disentanglement of factors of variations, which would be desired for counteracting imbalance for under-represented  factors and b) are only applicable when factors are  known a-priori. Our work leverages generative models (InfoStyleGAN) that address both the needs for analysis and synthesis, i.e. they discover de-novo, and then control for, unknown factors of variation within  data that may affect robustness, understood here as invariance / insensitivity of the performance of models vis-a-vis variations in such factors.
{\bf In-Processing Adversarial Approaches} intervene on the model to annihilate spurious learned correlations between factors’ and task labels~\cite{ganin2016domain}. This requirement drives at the heart the desire for conditional independence of predicted labels on the sensitive factor, which underpins many of the definitions of fair models. Many approaches offer variations on that strategy using adversarial two player settings, including \cite{beutel2017data,wadsworth2018achieving,zhang2018mitigating,alvi2018turning,song2019learning}. We also use this type of intervention here to render the model insensitive to unknown factors, but we complement it with a novel constraint, that assumes a new inductive bias regarding the semantic consistency of predicted labels vis-à-vis sensitive factor, which we call ``coherence regularization’’. {\bf Post-Processing  Approaches} usually intervene the model's predictions in some way, e..g via re-thresholding \cite{hardt2016}, or equalized calibration \cite{pleiss2017fairness}. Although those methods are useful, and particularly for tabular data, in general, their main limitation is in how much they can de-bias as they do not affect the model or data directly.  \cite{hardt2016} has shown evidence that separating debiasing from training is sub-optimal, especially for the deeper architectures used in computer vision.

{\bf Counterfactuals:}
Another line of work related to ours consists in using generative models that modifying real images in order to have certain factors changed. These modified images are commonly called counterfactuals, and differences in predictions between these counterfactuals and the original images are called {\em counterfactual fairness} \cite{kusner2017counterfactual}. In addressing counterfactual fairness, \cite{joo2020gender} uses an encoder-decoder architecture that takes in specific factors as input to produce counterfactuals to test public facial classification APIs. \cite{denton2019detecting} uses  labeled images to learn  linear classifiers in latent space for each factor, and consequently  perturb  latent codes to create counterfactuals. \cite{pawlowski2020deep} develops structured causal models for modeling using knowledge of what  factors affect, using generative models for specific operations in the model. \cite{dash2020evaluating} learns a structured causal models between the given factors, which is then used alongside an encoder-decoder architecture to produce counterfactuals. Although our goals are related to all these works  in trying to evaluate and mitigate bias depending on factors, all these assume a supervised setting where those factors are known a-priori. By contrast our work also can be seen as addressing counterfactual fairness, but we assume no a-prior knowledge  about factors. 

{\bf Analysis/Synthesis vis-a-vis Factors via Generative Models:}
Such models aim to align individual factors inherent in the data with latent space variables, in addition to the usual realism/diversity objectives. A major challenge is to ensure that intervening on a single factor leaves others invariant, a problem known as {\it disentanglement}.  
The effect was first noted in variational autoencoders (VAEs), and then subsequently in beta-VAE~\cite{wadsworth2018achieving}. \cite{jha2018disentangling} takes latent codes that should represent factors from real images and ensures they can be reconstructed from the image. PCA is used in  latent space of GANs in \cite{wang2019balanced}.
In
\cite{paul2020unsupervised}  the generator is forced to make the effect of certain latent variables align with the image variations; it  shows that optimizing for such  control of semantic factors also relates to {\it disentanglement} among latent factors. Interestingly, disentanglement is shown empirically a virtuous  effect on fairness~\cite{Locatello2019OnTF} but this work does not provide a method to achieve disentanglement. Our work follows in the foot steps of \cite{paul2020unsupervised} but is concerned with the problem of generalizing fairness (not an goal of that study).

In sum, our contributions are the following:
\begin{enumerate}
    \item {\em Fairness generalization and discovery of unknown sensitive factors of variation} We expand the scope of fairness and counterfactual studies focusing on solely known protected factors to more general settings of arbitrary factors, and importantly, we consider these factors are not known a-priori and  use generative models to discover them.
    \item {\em Unsupervised / Generalization / Semi-Supervised Settings} We consider different settings corresponding to different degrees of knowledge available about sensitive factors. We also address novel settings where no knowledge (unsupervised), some knowledge (semi-supervised) and ask the question how discovered factors may generalize from source to target factors. 
    \item {\em Intervention via ACAI} We evaluate the efficacy of achieving robustness and fairness vis-a-vis those factors using a triple set of interventions including augmentation,  coherence regularization (a novel intervention) and adversarial (ACAI) interventions. 
\end{enumerate}

\section{Methods}
We turn to definitions first, and counterfactual fairness, which is a concept we extends to unknown sensitive factors:

\begin{figure*}
\centering 
\includegraphics[page=2,width=14cm]{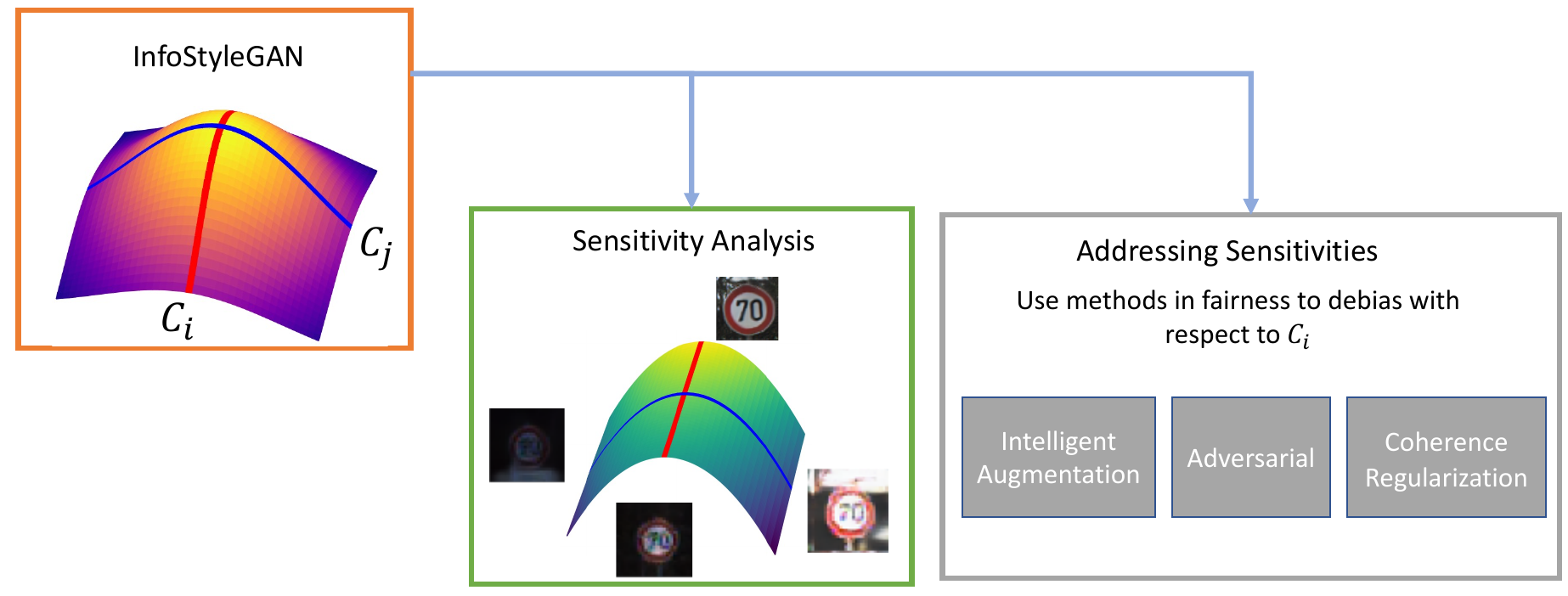}
     \vspace{-0.01in}
    \caption{We start with a generative model that is able to discover and control the desired factors. The specific generative model we use, InfoStyleGAN, uses mutual information to find latent space variables aligned with factors of variations of the data, similar to how PCA finds directions that explain the most variance. We can then use the generative model as a source of data to conduct a sensitivity analysis on the baseline model, and address these sensitivities using one of three methods.}
\label{fig:pipeline}
\end{figure*}

\subsection{Fairness without Labels}
Our approach to fairness without labels is related to counterfactual fairness \cite{kusner2017counterfactual}. Counterfactual fairness effectively asks whether a given factor $A$ is a true cause of changes in predictions for a predicted task label $\hat{Y}=g(X)$, with $X$ the input and $Y$ its actual task label. Commonly, $A$ and $X$ are said to be observable, or variables that are explicitly given. 
In causal inference, a causal model can be constructed incorporating $A$ and $X$ as well as latent background variables that are not given but are determined from a given $A$ and $X$. Using this model allows one to 'intervene' on $A$ corresponding to some $X$ and seeing how $X$ is affected. Counterfactual fairness is then the case where:
%


\begin{align}
    &P(\hat{Y}(\hat{X}_{A\leftarrow a} ) = y | X = x, A = a) = \nonumber \\
    &P(\hat{Y}(\hat{X}_{A\leftarrow a'}) = y | X = x, A = a)
\end{align}


\noindent where $\hat{X}_{A\leftarrow a'}$ denotes the intervention by setting $A$ to $a'$ in the causal model.

In contrast to the above definition, for our work, the only observable variable is $X$, making it difficult to evaluate causes for disparities in the predictor. In this work, we proceed by positing the existence of yet unknown  factors $\{A_i\}$ present in $X$ that could influence the behavior of the predictor. Consequently, the goal of this work is to recover these $\{A_i\}$ in variables $\{C_i\}$ and evaluate the fairness with respect to each $C_i$. One caveat is that exactly recovering some $A_i$ in a $C_i$ is difficult if not impossible without strong inductive biases \cite{Locatello2019OnTF}. In practice, the developer of the predictor can take the discovered $C_i$ as something that could plausibly correspond to a true sensitive factor $A_i$, and intervene on $C_i$.

Once the fairness has been evaluated with respect to these $C_i$, we can then mitigate any biases using  intervention methods such as discussed in prior work, such as augmenting the dataset, using an adversarial intervention on the model, or a technique called coherence regularization, inspired from a different application \cite{dash2020evaluating}. Our overall pipeline is therefore a two-pronged approach, first to discover factors of variations, then performing interventions, and is summarized in Figure \ref{fig:pipeline}. 

Next, we discuss step 1 in this dual-pronged method, i.e., how we discover $C_i$.

\subsection{Discovering Semantic Inherent Factors}

In order to recover $\{A_i\}$ in variables $\{C_i\}$, we follow the approach in \cite{paul2020unsupervised}. This technique builds on  \cite{zhao2017infovae} by augmenting latent code used by the generator with $\{C_i\}$ and using a StyleGAN architecture for the generator. Consequently, each individual $C_i$ can now influence the image generation at multiple scales, allowing for better control over potential attributes in the synthetic image. In order to ensure $C_i$ aligns with potential $A_i$, we require that $C_i$ be reconstructed from the image, where a network $Q$ predicts $C_i$ from the image. We take $Q$ to share weights with the discriminator $D$, where $Q$ takes as input an intermediate representation from $D$. To ensure that $C_i$ do not overlap with $Y$, we also make the generator conditional on $Y$. The loss function:

\begin{align}
L_{\text{info}} = \mathbb{E}_{\hat{X} \sim G(Z, C, Y)}(\mathbb{CE}(p(C|\hat{X}), Q(\hat{X})) - \log p(C))
\end{align}

\noindent is then added to the loss function for $G$ and $Q$, where $\mathbb{E}(\cdot)$ denotes the expectation operator and $\mathbb{CE}(\cdot;\cdot)$ the cross entropy.

Consequently, the full optimization scheme for the GAN is:

\begin{align}
    \min_{G, Q} \max_D \mathbb{E}_{X}{\log D(X)} + \mathbb{E}_{\hat{X} \sim G(\cdot)}(\log (1 - D(\hat{X}))+ L_{\text{info}}
\end{align}
We next turn to step 2 in our two-pronged approach: mitigating sensitivity to these $C_i$.

\subsection{Mitigation of Sensitivities}
\subsubsection{Semantic Augmentation (SA)}
For this intervention, we sample an additional fixed size dataset from our generator. We then append this data as supervised examples to the original dataset, and treat them as real data for the purposes of training the classifier. To target debiasing a specific factor $C_i$, we sample $C_i$ from a uniform distribution from -2 to 2 rather than the original distribution used in training the GAN. This uniform distribution was chosen to ensure that all possible values of the semantic factor are equally represented in the additional data. The parameters were chosen to overlap with the majority of the original Gaussian distribution, but excludes outliers that are less realistic similarly to the truncation trick \cite{karras2019style}.

\subsubsection{Adversarial Debiasing (AD)}

The previous method can be viewed as a passive intervention on the classifier training, and does not explicitly leverage the information provided by knowing what specific factor the generated image has. Thus, we also use an adversary to predict the sensitive factor (our $C_i$) and have the classifier defend against it as done in other papers such as \cite{zhang2018mitigating}.

During training, we structure our predictor $\hat{Y}$ as two consecutive components $\hat{Y}_1$ and $\hat{Y}_2$:
\begin{align}
\hat{Y}(X) = \hat{Y}_2(\hat{Y}_1 (X)) = \hat{Y}_2(R)
\end{align}

and construct an adversarial network $F$ that aims to predict the factor of variation $C_i$ from $R$. $R$ should then be invariant to any influences from $C_i$.

Simultaneously, the adversarial network ingests this internal representation $R$ and the true $Y$ to produce $\hat{C_i}$ (the prediction for $S$):
\begin{align}
F(R, Y) = \hat{C_i}
\end{align}
The cross entropy $\mathbb{CE}(\cdot;\cdot)$ is then applied to each of the two predictions and combined to compute the total loss which is optimized as follows:
\begin{equation}
    \min_{\hat{Y}} \max_{F} \mathbb{CE}(p(Y|X); \hat{Y}(X) ) - \beta \mathbb{CE}(p(C_i|\hat{X}); F(R, Y))
    \label{adv}
\end{equation}
where the $\beta>0$ is a hyperparameter used to balance the impact of the adversarial loss contribution. Combining the impact of the two loss terms ensures that the prediction network will be penalized for producing an $R$ that can be used to recover the protected factor (yielding a censoring effect). The resulting prediction network should then be more resilient to bias with respect to the protected factor.

\subsubsection{Coherence Regularization (CR)}

As an alternative to actively defending against an adversary predicting the sensitive factor, we can instead take as inductive bias  the view that the classifier should output the same probabilities regardless of the factor values. For example, if the learnt factor corresponds were to correspond to a known factor, say skin color, for a face verification task, then changing the skin color of a light skinned person to be darker ought not change the predictions of the classifier. Taking inspiration from \cite{xie2020unsupervised} and \cite{dash2020evaluating}, we enforce this invariance/insensitivity by adding a KL divergence between $\hat{Y}(G_{C_i \leftarrow c_i}(Z, C, Y)) = \hat{Y}(\hat{X}_{C_i \leftarrow c_i})$  and $\hat{Y}(G_{C_i \leftarrow c_i'}(Z, C, Y)) = \hat{Y}(\hat{X}_{C_i \leftarrow c_i'})$. Note that $\hat{X}_{C_i \leftarrow c_i'}$ denotes taking the (Z,C) that produced $\hat{X}$, setting $C_i$ to $c_i'$ and generating the corresponding image.
Thus, the full loss term for the classifier $F$ becomes:
\begin{align}
    L(F) &= \mathbb{E}_{X,Y} \mathbb{CE}(p(Y|X), F(X)) + \nonumber \\
    &\mathbb{E}_{\hat{X}\sim G(z,c,y)}\mathbb{KL}(\hat{Y}(\hat{X}_{C_i \leftarrow c_i}) || \hat{Y}(\hat{X}_{C_i \leftarrow c_i'})))
\end{align}

Note that unlike previous methods, the generator is used in an online fashion to generate the original synthetic examples and the perturbed example. Using a fixed synthetic dataset leads to a different formulation for what the expectation is over, in much the same way data augmentation is done in an online fashion rather than sampling a few specific augmentations.

\section{Settings} 
We consider three settings for evaluating outcomes:
\begin{enumerate}
\item {\bf Unsupervised}: this is the native setting of this work where we only assume we have  images and  labels for the task, but no knowledge on the sensitive factors, forcing us to discover sensitive factors and intervene on these. We refer to the goal in this setting as ({\bf Unsupervised Fairness}).
\item {\bf Generalization} in this setting we pursue another type of evaluation, asking the question if  interventions performed via our two pronged unsupervised approach are able to debias real factors. This tests a case of ``domain generalization'' where the change of domain is going from source/discovered factors to target/real factors. We call achieving this aim a type of ({\bf Fairness Generalization}).
\item {\bf Semi-supervised}: we consider here that additional information is made available for real factors and their corresponding labels which can be used effectively as a validation dataset to perform an automated selection of hyperparameters consisting of the choice of the discovered factor, combined with the selection of the intervention, that is best apt at debiasing for the disclosed validation factors. We call this aim a type of ({\bf semi-supervised}) adaptation from source/discovered to target/real factors.  
\end{enumerate}
\section{Experiments}
\subsection{Datasets}

We evaluate on two image datasets, German Traffic Sign Recognition Benchmark (GTSRB) \cite{Stallkamp-IJCNN-2011} and CelebA \cite{liu2015faceattributes}.

German Traffic Sign Recognition Benchmark is an image dataset containing 43 different classes of traffic signs. Common factors that are present in this dataset are primarily the size of the sign and the lighting condition, which can affect how finer details appear in the image. The image size is 64 pixels by 64 pixels, and we use 45,322 images for training and validation, and 10,112 images for testing. The synthetic dataset, correspondingly, is 101,120 images. The ground truth factors we take are the sign area, which is the area computed from the sign ROI normalized to be between 0 and 1, and the computed brightness, which is considered to be the luminance in CIELab color space averaged over the image.

CelebA is a facial image dataset consisting of celebrities, with many different attributes labeled ranging from the gender to the attractiveness of the person. We predict the age of the person. The images are cropped to be 128 pixels by 128 pixels, and we have 162,121 images for training, and 39,201 images for testing. The ground truth factors we take for this dataset are the skin color, which is the ITA computed over the face as in \cite{paul2020tara}, and the computed brightness of image, which is considered to be the luminance in CIELab color space averaged over the image.

The subpopulations for all ground truth factors except ITA for CelebA are taken to be one of 10 bins the factor is equally partitioned into. For ITA, there are only two subpopulations that are split using a threshold of 17, which was found in \cite{paul2020tara} to almost exclude facial images that denoted as 'Pale Skin'.

\subsection{Fairness Metrics}
We report overall accuracy to characterize utility. For fairness, we compute the accuracy gap over the subpopulations defined by $A$ or $C_i$ to be the maximum overall accuracy over the subpopulations minus the minimum overall accuracy:
\begin{align}
    \text{Acc}_{\text{gap}}(\hat{Y}) = \max_{j} \text{Acc}(\hat{Y}(\hat{X}_{A_{i}})) \nonumber \\ 
    - \min_{j} \text{Acc}(\hat{Y}(\hat{X}_{A_{j}}))
\end{align}
for a classifier $\hat{Y}$ where $A_i$ denotes a specific subpopulation. We also compute the minimum accuracy over the subpopulations, which aligns well with observing the worst case performance, and is a Rawlsian philosophy for fairness \cite{rawls2001justice}.

As an important question is the trade-off in fairness and utility, we also compute a compound metric $\text{CAI}_\lambda (\hat{Y})$ as in \cite{paul2020tara}, which is the weighted sum of the improvement in test accuracy between the classifier $\hat{Y}$ and the baseline and the decrease in accuracy gap from the baseline to $\hat{Y}$.
\begin{align}
    \text{CAI}_{\lambda}(\hat{Y}) = &\lambda (\text{Acc}_{\text{gap}}(\text{baseline}) - \text{Acc}_{\text{gap}}(\hat{Y})) + \nonumber \\
    &(1-\lambda ) (\text{Acc}(\hat{Y}) - \text{Acc}({\text{baseline}}))
\end{align}

\subsection{Experiments}
For each dataset, we first partition into a training dataset and a test dataset. The training dataset is used to both train InfoStyleGAN as well as the task classifiers. The test dataset is used to evaluate the overall fairness and utility metrics for every method, and is also used to compute fairness metrics in settings where test labels for ground truth factors are available.

We first train InfoStyleGAN on the dataset by dedicating ten $C_i$ variables, to discover sensitive factors innate in the data. We train a baseline task classifier, which is a ResNet50 using pre-trained ImageNet weights, on the given task for the dataset.  

For the {\bf unsupervised setting}, we  intervene on all sensitive factors for the baseline and evaluate the resulting change in fairness metrics. We evaluate  four models (baseline model and baseline intervened with the three mitigations we described in the methods section). This evaluation is done on synthetic images as follows.
To evaluate performance with respect to discovered factors for the synth factors setting, we sample data conditioned on the class labels from InfoStyleGAN where each $C_i$ is sampled uniformly from -2 to 2 for the same reasons as the augmentation method. We partition the resulting images according to the image's $C_i$, where there are 10 bins between -2 and 2. This synthetic dataset has the same label distribution as the test set, and is sampled to be ten times the size, so each partition is roughly equal size to the original test set. We then compute the overall accuracy for each image partition, and treat these individual partitions as subpopulations to compute the fairness metrics.

For the {\bf generalization setting} the high level goal is to assess how well an intervention developed under the unsupervised setting actually performs when generalized and applied to a specific / real sensitive factor. Therefore testing is here done for a factor that appears to align semantically with a discovered factor of variation. For this setting we report the resulting performance corresponding to all three types of interventions. 

For the {\bf semi-supervised setting}, we instead assume we now have access to a validation dataset of data with labels for the real sensitive factor, this data is now  used to perform hyperparameter selection (intervention and factor to intervene). In effect when we have the validation labels, we  intervene on every $C_i$ and use the validation labels to select the best pair of $C_i$ and intervention type. We denote the selected intervention as Intervention-i, e.g.  SA-1 denotes the semantic augmentation on $C_1$. 

All results for all settings are reported in the same tables for a given sensitive factor.

\begin{table}[t]
\includegraphics[width=\linewidth]{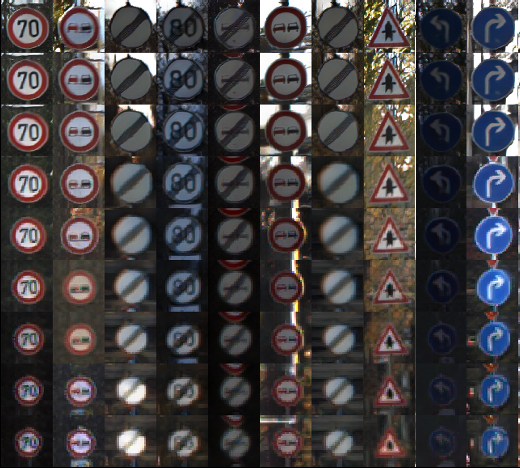}

\centering
\scriptsize
    \centering
    	\begin{tabular}{c|c|c|c|c|c|c}
	    \toprule
        Setting&Intervene&Acc&$\text{Acc}_{\text{gap}}$&$\text{Acc}_{\text{min}}$& $\text{CAI}_{0.5}$  &  $\text{CAI}_{0.75}$  \cr
        \midrule
        \midrule
        \multirow{4}{*}{\parbox{1.0cm}{\centering \textbf{Unsuperv. Fairness}}}&No &98.49&22.93&65.80&0.00&0.00 \cr
        &SA-1 &\textbf{98.66}&\textbf{4.12}&79.31&\textbf{9.49}&\textbf{14.17} \cr 
        &AD-1 &97.99&10.84&\textbf{86.09}&5.80&8.96 \cr
        
        &CR-1 &96.31&34.85&48.50&-7.05&-9.52 \cr \cline{1-7}
        \midrule
        \multirow{4}{*}{\parbox{1.0cm}{\centering \textbf{Fairness Generalization}}}&No &98.49&3.33&96.24&0.00&0.00 \cr
        &SA-1 &\textbf{98.66}&\textbf{2.83}&\textbf{96.82}&\textbf{0.34}&\textbf{0.42} \cr 
        &AD-1 &97.99&\textbf{2.83}&96.49&0.00&0.25 \cr
        
        &CR-1 &96.31&5.56&92.76&-2.20&-2.22 \cr \cline{1-7}
        \multirow{3}{*}{\parbox{1.0cm}{\centering \textbf{Semi-superv. Fairness}}}&\multirow{3}{*}{\parbox{0.8cm}{ACAI (SA-4)}} &\multirow{3}{*}{\textbf{99.12}}&\multirow{3}{*}{\textbf{1.42}}&\multirow{3}{*}{\textbf{98.15}}&\multirow{3}{*}{\textbf{1.27}}&\multirow{3}{*}{\textbf{1.59}} \cr
        &&&&&\cr
        &&&&&\cr
        \bottomrule
	\end{tabular}
	
    \captionof{table}{GTSRB Image transitions and results for Sign Size. Sensitive Factor 1 visually corresponds most to this factor. Fairness metrics for synth factor settings are computed with respect to factor 1, and fairness metrics for real test factors are computed with respect to the true sign area factor. If the real sensitive factor is available to the developer, we use it to select the best method and factor to intervene on.}
    \label{tab:gtsrb1}
\end{table}
\begin{table}[t]

\includegraphics[width=\linewidth]{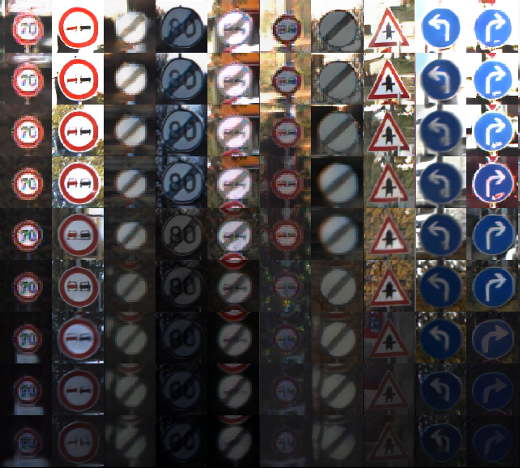}

\scriptsize
    \centering
    	\begin{tabular}{c|c|c|c|c|c|c}
	    \toprule
        Setting&Intervene&Acc&$\text{Acc}_{\text{gap}}$&$\text{Acc}_{\text{min}}$& $\text{CAI}_{0.5}$  &  $\text{CAI}_{0.75}$  \cr
        \midrule
        \midrule
        \multirow{4}{*}{\parbox{1.0cm}{\centering \textbf{Unsuperv. Fairness}}}&No &98.49&18.46&69.37&0.00&0.00 \cr
        &SA-7 &\textbf{99.06}&\textbf{2.81}&\textbf{95.82}&\textbf{8.09}&\textbf{11.87} \cr 
        &AD-7 &98.01&4.45&92.00&6.74&10.38 \cr
        
        &CR-7 &95.96&15.65&65.30&0.10&1.46 \cr \cline{1-7}
        \midrule
        \multirow{4}{*}{\parbox{1.0cm}{\centering \textbf{Fairness Generalization}}}&No &98.49&\textbf{2.10}&\textbf{97.37}&\textbf{0.00}&\textbf{0.00} \cr
        &SA-7 &\textbf{99.06}&3.32&96.50&-0.33&-0.78 \cr 
        &AD-7 &98.01&3.13&95.98&-0.75&-0.89 \cr
        
        &CR-7 &95.96&3.81&94.09&-2.12&-1.91 \cr \cline{1-7}
        \multirow{3}{*}{\parbox{1.0cm}{\centering \textbf{Semi-superv. Fairness}}}&\multirow{3}{*}{\parbox{0.8cm}{ACAI (SA-9)}} &\multirow{3}{*}{\textbf{98.93}}&\multirow{3}{*}{\textbf{1.67}}&\multirow{3}{*}{\textbf{97.81}}&\multirow{3}{*}{\textbf{0.44}}&\multirow{3}{*}{\textbf{0.43}} \cr
        &&&&&\cr
        &&&&&\cr
        \bottomrule
	\end{tabular}
	
    \captionof{table}{GTSRB Image transitions and results for lighting. Sensitive Factor 7 visually corresponds most to this factor as shown above.}
    
    \label{tab:gtsrb7}
\end{table}

\subsection{Results}
 We show the results of intervening on the two most sensitive factors for the baseline in the unsupervised setting for both the CelebA and the traffic sign experiment herein (the results for all ten factors are shown in the appendix).

We show those results in tables \ref{tab:gtsrb1}, \ref{tab:gtsrb7}, \ref{tab:celeba5}, and \ref{tab:celeba4}. Each table has a figure above it which corresponds to intervening only on the specified factor in the generator on the y-axis, and each column corresponds to a individual sample taken.  For the table itself, each numerical column corresponds to a specific metric used, where the fairness metrics for unsupervised fairness are computed over $C_i$ and the fairness metrics for the other two settings are over $A$. 

\section{Discussion}
\subsection{Observations}
\subsubsection{Unsupervised Fairness}
For this setting, we see that the initial high accuracy gaps for the model with no intervention are mitigated significantly by intervening on the sensitive factor. Typically, intervening also slightly improved the overall accuracy on real data, except for Table \ref{tab:celeba5} for lighting, where the highest accuracy over the interventions is 80.38\% for CR-5. Due to this, the methods that the CAI metrics were highest at were also the ones that had the lowest accuracy gap.

Semantic Augmentation was the best performing intervention for all metrics except for Table \ref{tab:celeba4} where the coherence regularization has a better overall accuracy and $\text{CAI}_\text{0.5}$. For the GTSRB dataset, the AD method came in second compared to semantic augmentation and still improved over the baseline. Coherence Regularization is worse than the other two interventions, and worse than the baseline when addressing sign size. For CelebA, adversarial debiasing exhibited worse overall accuracies compared to coherence regularization. Coherence regularization had thesecond best accuracy gap for skin color, and the second best accuracy and second worst accuracy gap for lighting.
\subsubsection{Fairness Generalization}
For GTSRB, we see that the SA and AD interventions improve upon the classifier with no intervention for sign area in terms of both the accuracy gap as well as the minimum accuracy. However, no intervention is best in terms of fairness for lighting in this setting. Of the interventions, semantic augmentation performed best in terms of $\text{CAI}_{0.5}$.

For CelebA, coherence regularization had the best $\text{CAI}_{0.5}$, having the best accuracy gap for lighting and best overall accuracy for skin color. Only coherence regularization improved upon the original model for lighting, but all three interventions improved in either overall accuracy or the accuracy gap for skin color.
\subsubsection{Semi-supervised Fairness}
Due to intervening over any factor and approach, the ACAI methods have the highest $\text{CAI}_{0.5}$ over all interventions evaluated on real factors. For GTSRB, ACAI primarily improved the accuracy gap, whereas for CelebA ACAI has a slight decrease in the accuracy gap in return for an increase in the overall accuracy.
\subsection{Limitations}
\begin{table}[]

\includegraphics[width=\linewidth]{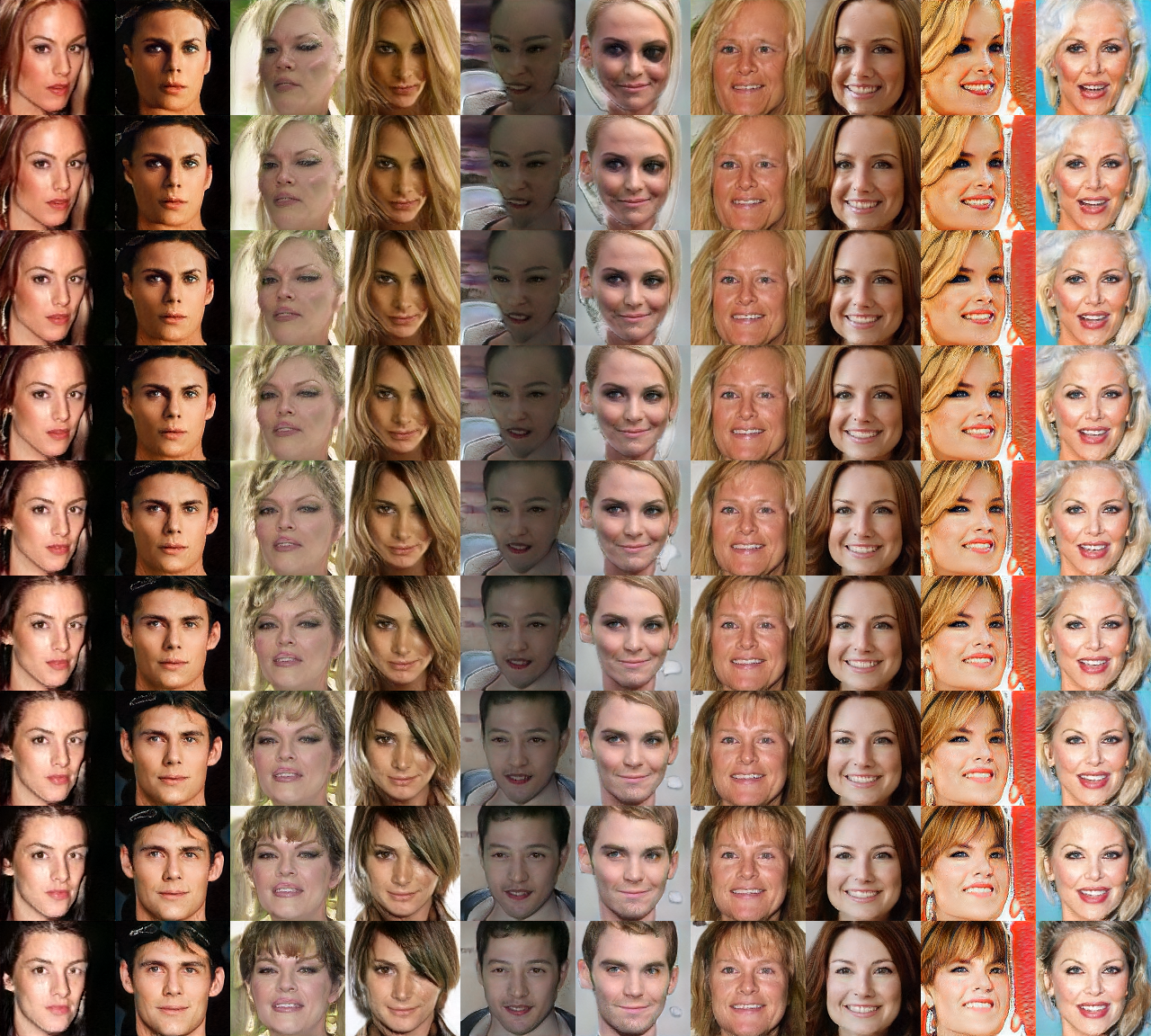}

\scriptsize
    \centering
    	\begin{tabular}{c|c|c|c|c|c|c}
	    \toprule
        Setting&Intervene&Acc&$\text{Acc}_{\text{gap}}$&$\text{Acc}_{\text{min}}$& $\text{CAI}_{0.5}$  &  $\text{CAI}_{0.75}$  \cr
        \midrule
        \midrule
        \multirow{4}{*}{\parbox{1.0cm}{\centering \textbf{Unsuperv. Fairness}}}&No &\textbf{81.34}&8.77&69.80&0.00&0.00 \cr
        &SA-5 &79.45&\textbf{1.02}&\textbf{86.34}&\textbf{2.93}&\textbf{5.30} \cr 
        &AD-5 &74.11&1.36&84.37&0.15&3.79 \cr
        
        &CR-5 &80.38&7.86&67.94&0.02&0.47 \cr \cline{1-7}
        \midrule
        \multirow{4}{*}{\parbox{1.0cm}{\centering \textbf{Fairness Generalization}}}&No &\textbf{81.34}&6.07&78.20&0.00&0.00 \cr
        &SA-5 &79.45&6.86&\textbf{79.93}&-1.34&-1.06 \cr 
        &AD-5 &74.11&10.41&67.96&-5.78&-5.06 \cr
        
        &CR-5 &80.38&\textbf{4.64}&78.08&\textbf{0.23}&\textbf{0.83} \cr \cline{1-7}
        \multirow{3}{*}{\parbox{1.0cm}{\centering \textbf{Semi-superv. Fairness}}}&\multirow{3}{*}{\parbox{0.8cm}{ACAI (SA-2)}} &\multirow{3}{*}{\textbf{82.53}}&\multirow{3}{*}{\textbf{5.41}}&\multirow{3}{*}{\textbf{79.46}}&\multirow{3}{*}{\textbf{0.92}}&\multirow{3}{*}{\textbf{0.79}} \cr
        &&&&&\cr
        &&&&&\cr
        \bottomrule
	\end{tabular}
	
    \captionof{table}{CelebA Image transitions and results for sensitive factor 5 corresponding to lighting. Fairness metrics for synth factor settings are computed with respect to factor 5, and fairness metrics for real test factors are computed with respect to the true sign area factor. If the real sensitive factor is available to the developer, we use it to select the best method and factor to intervene on.}
    \label{tab:celeba5}
\end{table}
\begin{table}[]

\includegraphics[width=\linewidth]{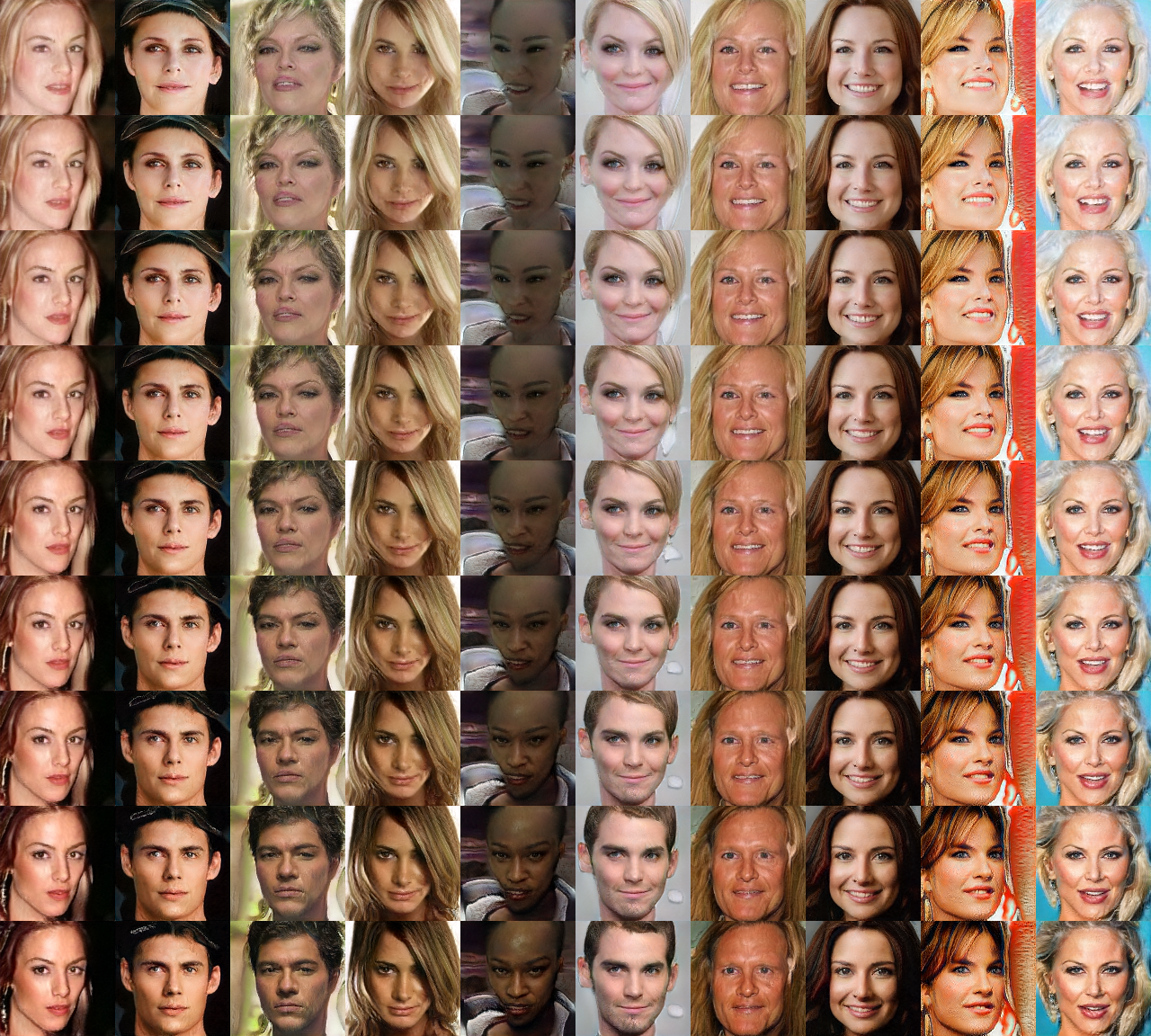}

\scriptsize
    \centering
    	\begin{tabular}{c|c|c|c|c|c|c}
	    \toprule
        Setting&Intervene&Acc&$\text{Acc}_{\text{gap}}$&$\text{Acc}_{\text{min}}$& $\text{CAI}_{0.5}$  &  $\text{CAI}_{0.75}$  \cr
        \midrule
        \midrule
        \multirow{4}{*}{\parbox{1.0cm}{\centering \textbf{Unsuperv. Fairness}}}&No &81.34&7.20&70.09&0.00&0.00 \cr
        &SA-4 &\textbf{82.53}&\textbf{1.05}&\textbf{86.19}&\textbf{3.72}&\textbf{4.92} \cr 
        &AD-4 &80.12&3.99&81.59&0.99&2.10 \cr
        
        &CR-4 &81.48&2.12&69.48&2.58&3.83 \cr \cline{1-7}
        \midrule
        \multirow{4}{*}{\parbox{1.0cm}{\centering \textbf{Fairness Generalization}}}&No &81.34&5.27&76.83&0.00&0.00 \cr
        &SA-4 &\textbf{82.53}&5.32&77.96&0.57&0.26 \cr 
        &AD-4 &80.12&\textbf{2.86}&77.67&0.59&\textbf{1.50} \cr
        
        &CR-4 &81.48&3.96&\textbf{78.09}&\textbf{0.72}&1.02 \cr \cline{1-7}
        \multirow{3}{*}{\parbox{1.0cm}{\centering \textbf{Semi-superv. Fairness}}}&\multirow{3}{*}{\parbox{0.8cm}{ACAI (CR-1)}} &\multirow{3}{*}{\textbf{81.94}}&\multirow{3}{*}{\textbf{3.07}}&\multirow{3}{*}{\textbf{79.31}}&\multirow{3}{*}{\textbf{1.40}}&\multirow{3}{*}{\textbf{1.80}} \cr
        &&&&&\cr
        &&&&&\cr
        \bottomrule
	\end{tabular}
    \captionof{table}{CelebA Image transitions and results for skin color. Sensitive Factor 4 visually corresponds most to this factor as shown above.}
    \label{tab:celeba4}
\end{table}
The generative models used were able to control factors within each domain while typically leaving other factors unaffected. How exactly the model chooses the factors it can control is unknown, and being able to specify the exact number and structure of these learnt factors is difficult at best. The model we use, and most approaches for discovering factors, focus on more 'global' factors that affect the image completely, as we see in the factors our InfoStyleGAN models capture. More complex scenes such as those commonly found in autonomous driving or aerial imagery have 'local' factors that affect only single entities in scenes, such as vehicles or people. To the best of our knowledge, capturing local factors has not been researched extensively.

Another issue is how consistent and diverse the factors we do capture are. Discovering factors is typically done in conjunction with a generative model, so this process of discovery can be thought of as how the model organizes itself to do the task. Latent representations in VAEs need to reconstruct the image exactly, so all important factors should be represented in some capacity. However, there may be no easy way to actually control them, as VAEs encourage representations that are sparse and have variables overloaded with multiple factors. Approaches using GANs, like this work employs, are the opposite in that the control is more consistent, but they may only capture a subset of important factors. Even then, there are effects on other factors as seen for CelebA associated with changing a factor that may be too much for other fairness methods such as the adversarial or coherence regularization we employ. Consequently, we view this work as an alternative to data collection rather than data curation, and needing to verify if the data or labels is of sufficient quality is still necessary. In line with this concern is claiming a classifier is fair only using the unsupervised setting without heavily vetting the controlled factor or evaluating on real data as in fairness generalization.

Longer term, as these generative models are developed and better describe the true domain via these latent factors, there is a natural question if whether is the generative model by itself is sufficient to do the task, i.e., there is no real need to train another classifier.  After all, there are generative classifiers such as naive Bayes, and even though GANs do not estimate the probability of an image directly one could plausibly use surrogate measures such as perceptual distance with respect to latent space.  However, these generative models may need to be significantly large to properly model the domain, making it intractable for most scenarios. 
Consequently, the view that we take is that our work has similar goals as that in distillation. Rather than trying to directly replace a model with something simpler, we instead want to take our knowledge that these factors should not be relied on in order to do the task, and intervene on the data. Being able to make this connection more directly is a potential future direction for this work.

\section{Conclusions}

We demonstrate that methods that can control factors operate harmoniously in the fairness regime. Using these generative models allows one to supplement data collection, and are useful when one does not significant amounts of labeled data to use for fairness. Finally, we note that some fairness methods may have challenges in dealing with noisy labels, showing a need for quality curation of the dataset.

{\small
\bibliographystyle{ieee_fullname}
\bibliography{egbib}
}

\end{document}